\documentclass[final]{elsarticle}
\usepackage{hyperref}
\usepackage{url}
\usepackage{booktabs}       
\usepackage{amsfonts}       
\usepackage{nicefrac}       
\usepackage{microtype}      
\usepackage{graphicx}
\usepackage{amsmath}
\usepackage{multirow}
\usepackage[utf8]{inputenc} 
\usepackage[T1]{fontenc}    
\usepackage{lineno}
\usepackage{color}
\modulolinenumbers[5]

\journal{Pattern Recognition}

\begin{document}

\begin{frontmatter}

\title{Relevance Attack on Detectors}


\author[mymainaddress]{Sizhe Chen}
\ead{sizhe.chen@sjtu.edu.cn}

\author[mymainaddress]{Fan He}
\ead{hf-inspire@sjtu.edu.cn}

\author[mymainaddress]{Xiaolin Huang\corref{mycorrespondingauthor}}
\cortext[mycorrespondingauthor]{Corresponding author}
\ead{xiaolinhuang@sjtu.edu.cn}

\author[mysecondaryaddress]{Kun Zhang}
\ead{kunz1@cmu.edu}

\address[mymainaddress]{Department of Automation, the Institute of Medical Robotics, and the MOE Key Laboratory of System Control and Information Processing, Shanghai Jiao Tong University, 800 Dongchuan Road, Shanghai, 200240, P.R. China.}
\address[mysecondaryaddress]{Philosophy Department, and Machine Learning Department, Carnegie Mellon University, 5000 Forbes Ave, Pittsburgh, PA 15213, United States}

\begin{abstract}
This paper focuses on high-transferable adversarial attacks on detectors, which are hard to attack in a black-box manner, because of their multiple-output characteristics and the diversity across architectures. To pursue a high attack transferability, one plausible way is to find a common property across detectors, which facilitates the discovery of common weaknesses. We are the first to suggest that the relevance map from interpreters for detectors is such a property. Based on it, we design a Relevance Attack on Detectors (RAD), which achieves a state-of-the-art transferability, exceeding existing results by above 20\%. On MS COCO, the detection mAPs for all 8 black-box architectures are more than halved and the segmentation mAPs are also significantly influenced. Given the great transferability of RAD, we generate the first adversarial dataset for object detection and instance segmentation, i.e., Adversarial Objects in COntext (AOCO), which helps to quickly evaluate and improve the robustness of detectors.
\end{abstract}

\begin{keyword}
adversarial attack, attack transferability, black-box attack, relevance map, interpreters, object detection.
\end{keyword}

\end{frontmatter}


\section{Introduction}
Adversarial attacks \cite{szegedy2013intriguing, goodfellow2014explaining, carlini2017towards, madry2017towards, su2019one, 8807315, 9072347, 9063523, 8725541, 8423654, 9025211, GHOSH2022108279} have revealed the fragility of Deep Neural Networks (DNNs) by fooling them with elaborately-crafted imperceptible perturbations. Among them, the black-box attack, i.e., attacking without knowledge of their inner structure and weights, is much harder, more aggressive, and closer to real-world scenarios. For classifiers, there exist some promising black-box attacks \cite{dong2018boosting, xie2019improving, lin2019nesterov, 9238430}. It is also severe to attack object detection \cite{zhang2019towards} in a black-box manner, e.g., hiding certain objects from unknown detectors \cite{thys2019fooling}. By that, life-concerning systems based on detection such as autonomous driving and security surveillance could be hurt when the black-box attack is conducted physically \cite{huang2020universal, Wang_2021_CVPR}.

To the best of our knowledge, no existing attack is specifically designed for black-box transferability in detectors, because they have multiple-outputs and a high diversity across architectures. In such situations, adversarial samples do not transfer well  \cite{su2018robustness}, and most imperceivable attacks only decrease mAP of black-box detectors by 5 to 10\% \cite{xie2017adversarial, li2018robust, li2018exploring}. To overcome this, we propose one feasible way to find common properties across detectors inspired by our AoA attack \cite{9238430}, which facilitates the discovery of common weaknesses. Based on them, the designed attack can threaten various victims.

\begin{figure}[!htpb]
  \centering
  \includegraphics[width=\hsize]{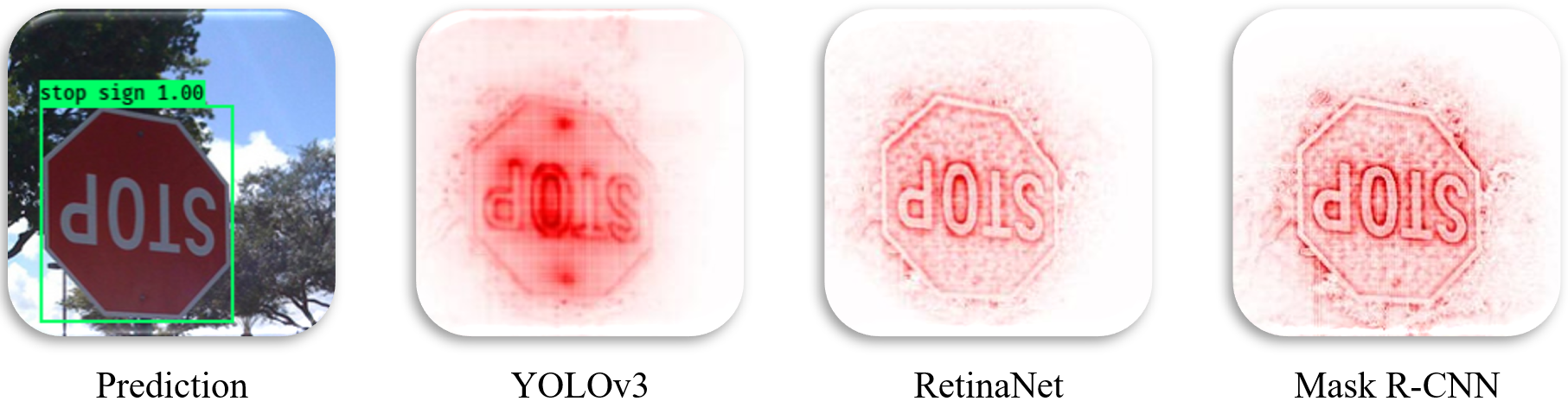}\\
  \caption{Relevance maps for models with different architectures. Three models not only predict the ``stop sign'' right, but also share similar relevance maps.}
  \label{intro}
\end{figure}

In this paper, we adopt the relevance map from DNN interpreters as common property, on which different detectors have similar interpretable results, as shown in Fig. \ref{intro} and \ref{visualrlv}. Based on relevance maps, we design a Relevance Attack on Detectors (RAD). RAD focuses on suppressing the relevance map rather than directly attacking the prediction as in existing works \cite{xie2017adversarial, li2018robust, li2018attacking, LI2021107584, XIAO2021107903}. Because the relevance maps are quite similar across models, those of black-box models are influenced and misled as well in attack, leading to great transferability. Although some works have adopted the relevance map as an indicator or reference of success attacks \cite{9238430, dong2019evading, zhang2019interpreting, wu2020boosting}, there is no work to directly attack the relevance maps of detectors to the best of our knowledge.

In our comprehensive evaluation, RAD achieves the state-of-the-art transferability on 8 black-box models for COCO) dataset \cite{lin2014microsoft}, nearly halving the detection mAP under the common $\ell_\infty$-bounded settings, and impairing detectors' performance in three sub-tasks. Interestingly, the adversarial samples of RAD also greatly influence the performance of instance segmentation, even only detectors are attacked. Given the high transferability of RAD, we create Adversarial Objects in COntext (AOCO), the first adversarial dataset for object detection and instance segmentation. AOCO contains 10K samples that significantly decrease the performance of black-box models for detection and segmentation. AOCO may serve as a benchmark to test the robustness of a DNN or improve it by adversarial training. To reproduce our results and access our dataset, one could visit \url{https://github.com/AllenChen1998/RAD}.

\subsection*{Contributions}
\begin{itemize}
  \item We propose a novel attack framework on relevance maps for detectors. We extend DNN interpreters to detectors, find out the most suitable outputs to attack by relevance maps, and explore the best update techniques to increase the transferability.
  \item We evaluate RAD on 8 black-box models and find its state-of-the-art transferability, which exceeds existing results by above 20\% in mAP. Detection and segmentation performance is greatly impaired in various metrics, invalidating the state-of-the-art DNN to a very rudimentary counterpart.
  \item By RAD, we create the first adversarial dataset for object detection and instance segmentation, i.e., AOCO. As a potential benchmark, AOCO is generated from COCO and contains 10K high-transferable samples. AOCO helps to quickly evaluate and improve the robustness of detectors.
\end{itemize}

\section{Related Work}\label{related}
Since \cite{szegedy2013intriguing}, there have been lots of promising adversarial attacks \cite{goodfellow2014explaining, carlini2017towards, madry2017towards}. Generally, they fix the network weights and change the input slightly to optimize the attack loss. The network then predicts incorrectly on adversarial samples with high confidence. \cite{papernot2017practical} find that adversarial samples crafted by attacking a white-box surrogate model may transfer to other black-box models as well. Input modification \cite{xie2019improving, dong2019evading, lin2019nesterov} or other optimization ways \cite{dong2018boosting, lin2019nesterov} are validated to be effective in enhancing the transferability.

\cite{xie2017adversarial} extends adversarial attacks to detectors. It proposes to attack on densely generated bounding boxes. After that, losses about localization and classification are designed \cite{li2018robust} for attacking detectors. \cite{lu2017adversarial} and \cite{li2018attacking} propose to attack detectors in a restricted area. Existing restricted digital attacks achieve good results in white-box scenarios but are not specifically designed for transferability. The adversarial impact on black-box models is quite limited, i.e., a 5 to 10\% decrease from the original mAP, even when two models only differ in backbone \cite{xie2017adversarial, li2018robust, li2018exploring}. \cite{wang2020adversarial} discusses black-box attacks towards detectors based on queries rather than the transferability as we do. The performance is satisfactory, but it requires over 30K queries, which is easy to be discovered by the model owner. Besides, physical attacks on white-box detectors are also feasible \cite{huang2019adversarial, wu2020making, xu2020adversarial}.

For great transferability, we propose to attack relevance maps, which are calculated by DNN interpreters \cite{zeiler2014visualizing, selvaraju2017grad, shrikumar2017learning, montavon2017explaining}. They are originally developed to interpret how DNNs predict and help users gain trust in them. Specifically, they display how the input contributes to a certain output in a pixel-wise manner. Typical works include Layer-wise Relevance Propagation (LRP) \cite{bach2015pixel}, Contrastive LRP \cite{gu2018understanding} and Softmax Gradient LRP (SGLRP) \cite{iwana2019explaining}. These methods encourage the reference of relevance maps in attack \cite{9238430, dong2019evading, zhang2019interpreting, wu2020boosting}, and also inspire us. However, none of them attack relevance maps for detectors.

RAD differs from \cite{ghorbani2019interpretation, zhang2020interpretable} in the goal. RAD misleads detectors by suppressing relevance maps. \cite{ghorbani2019interpretation} misleads the relevance maps while keeping the prediction unchanged. \cite{zhang2020interpretable} also misleads DNNs, but it keeps the relevance maps unchanged.

\section{Relevance Attack on Detectors}
We propose an attack specifically designed for black-box transferability, named Relevance Attack on Detectors (RAD). RAD suppresses multi-node relevance maps for several bounding boxes. Since the relevance map is commonly shared by different detectors as shown in Fig. \ref{intro}, attacking it in the white-box surrogate model achieves a high transferability towards black-box models. In this section, we first provide a high-level overview of RAD, and analyze the potential reasons for its transferability. Then we thoroughly discuss three crucial concrete issues in RAD.
\begin{itemize}
  \item In Section \ref{what}, we extend existing classifier interpreters to detector ones.
  \item In Section \ref{where}, we study the proper output scalars to attack by RAD.
  \item In Section \ref{how}, we explore the suitable techniques to update samples.
\end{itemize}

\subsection{What is RAD?}
\begin{figure}[!htpb]
  \centering
  \includegraphics[width=\hsize]{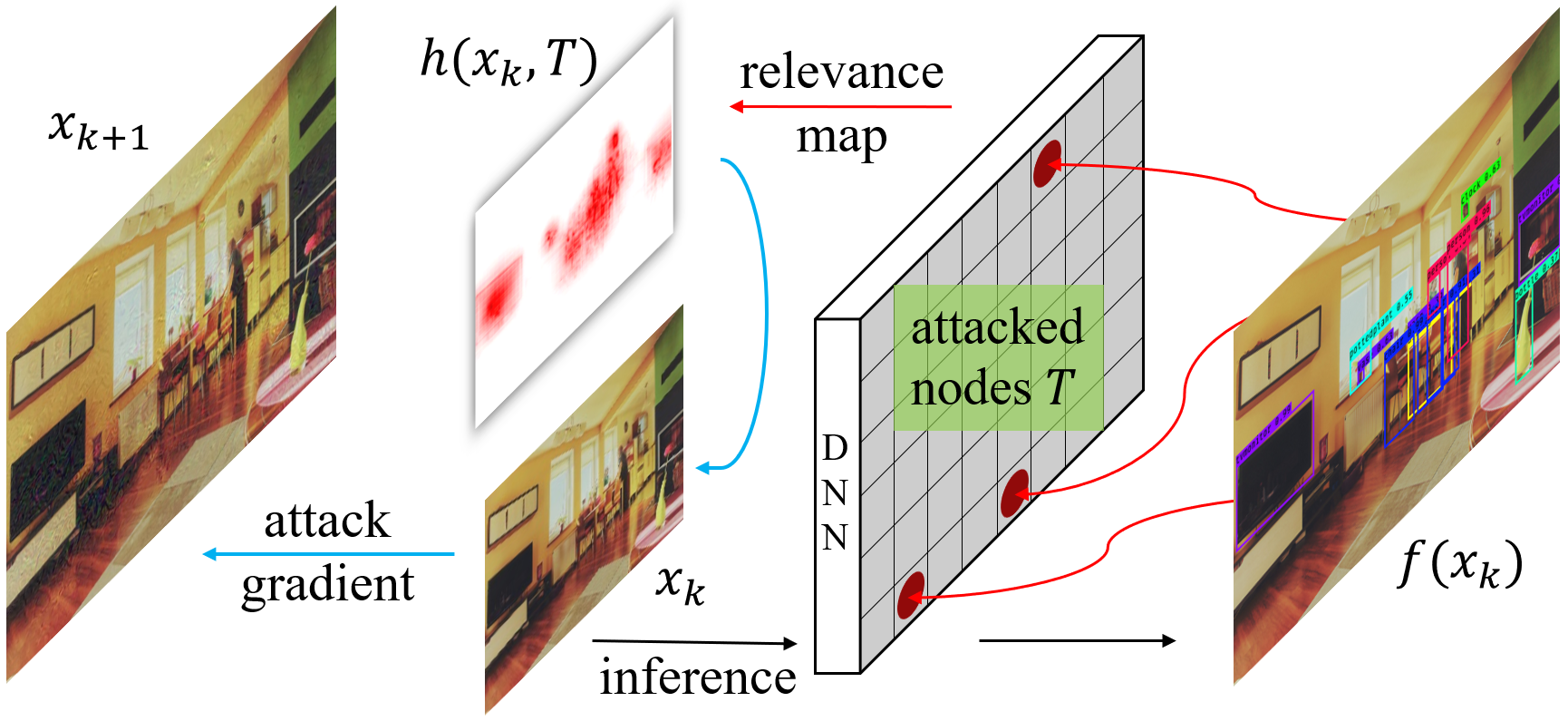}\\
  \caption{Framework of RAD. $x_k$ is the sample in iteration $k$ and $f(x_k)$ is the network prediction for it. $h(x_k, T)$ stands for the relevance map for all attacked nodes in $T$. RAD works by repeating processes denoted by ``black'', ``red'' and ``blue'' arrows in turn.}
  \label{architecture}
\end{figure}

We present the framework of RAD in Fig. \ref{architecture}. Initialized by the original sample $x_0$, the adversarial sample $x_k$ in the $k^\text{th}$ iteration is forward propagated in the surrogate model, getting the prediction $f(x_k)$. Current attacks generally suppress the prediction values of all attacked output nodes in $T$, where an output node stands for an output scalar of the detector. In contrast, RAD suppresses the corresponding relevance map $h(x_k, T)$. To restrain that, gradients of $h(x_k, T)$ back propagate to $x_k$, which is then modified to $x_{k+1}$.

Notably, RAD is a complete framework to attack detectors, and its components require special design. Besides the calculation of relevance maps of detectors, other components in RAD, e.g., the attacked nodes or the update techniques, also need a customized analysis. The reason is that no existing work directly attacks the relevance of detectors, and the experience in attacking predictions is not applicable here. For example, \cite{zhang2019towards} emphasizes classification loss and localization loss equally, but the former is validated to be significantly better in attacking the relevance in Section \ref{where}.

\subsection{Why RAD Transfers?}
\begin{figure}[!htpb]
  \centering
  \includegraphics[width=\hsize]{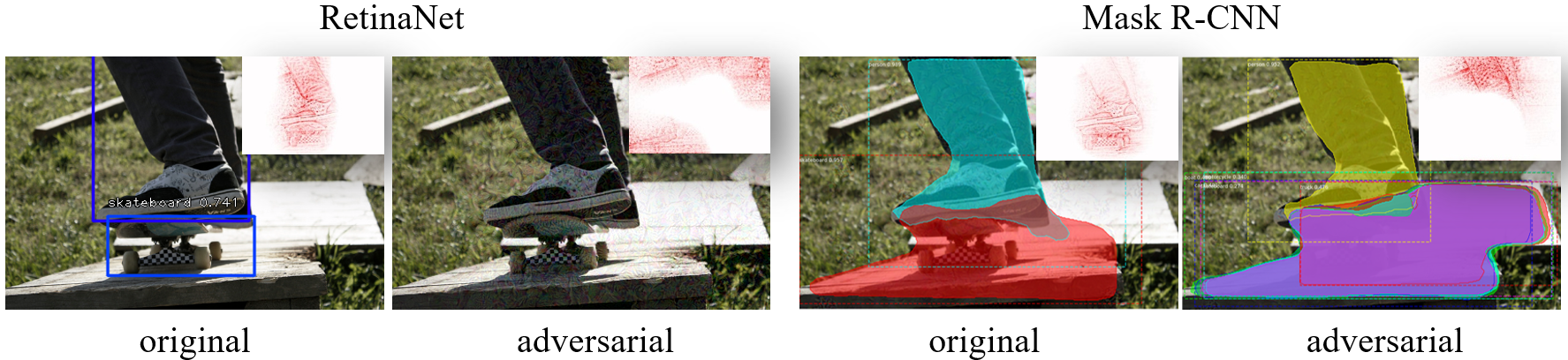}\\
  \caption{RAD's transferability origins from the change of relevance maps. The image contains a person and a skateboard. By attacking on relevance maps, both surrogate models make extremely confusing predictions.}
  \label{white}
\end{figure}

RAD's transferability comes from the attack goal: changing the common properties, i.e., the relevance maps, which are the same for different detectors because they are developed to interpret the salient parts in the data, and thus is data-dependent and model-independent \cite{simonyan2013deep, zeiler2014visualizing, selvaraju2017grad} for diverse well-trained models, which could be observed in Fig. \ref{intro} and \ref{visualrlv}. As shown in Fig. \ref{white}, the relevance maps are clear and structured for the original sample in both detectors. After RAD, the relevance maps are induced to be meaningless without a correct focus, leading to wrong predictions, i.e., no or false detection. Because relevance maps transfer well across models, those for black-box detectors are also significantly influenced, causing a great performance drop, which is illustrated visually in Section \ref{expvisual}.

\begin{figure}[!htpb]
  \centering
  \includegraphics[width=0.24\hsize]{imgs/p_sample_001532.png}
  \includegraphics[width=0.24\hsize]{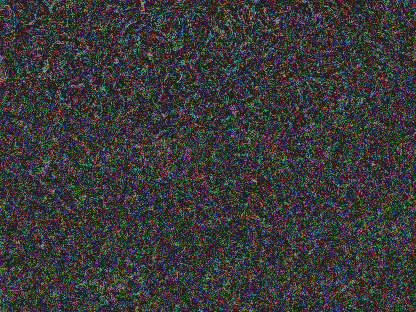}
  \includegraphics[width=0.24\hsize]{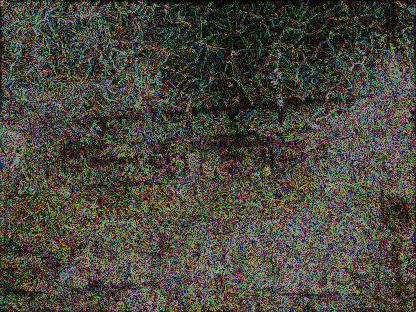}
  \includegraphics[width=0.24\hsize]{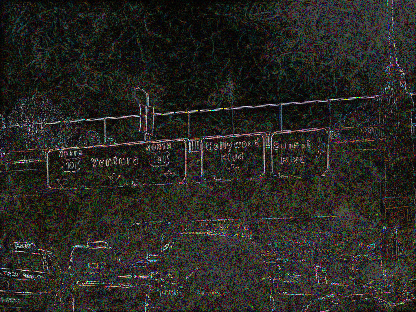}\\
  \caption{The original image and the adversarial perturbations ($\times 5$ in magnitude for demonstration) generated by Dfool \cite{lu2017adversarial}, DAG \cite{xie2017adversarial}, and RAD (from left to right)}
  \label{per}
\end{figure}

RAD also attacks quite ``precisely'', i.e., the perturbation pattern is significantly focused on distinct areas and has a clear structure as shown in Fig. \ref{per}. That is to say, RAD accurately locates the most discriminating parts of a sample and concentrates the perturbation on them, leading to a great transferability when the perturbations are equally bounded.

\subsection{What is the Relevance Maps for Detectors?}\label{what}
We analyze the potential of RAD above, below we make it feasible by addressing three crucial issues. To conduct the relevance attack, we first need to know the relevance maps for detectors.

Currently, there exist lots of interpreters to calculate the relevance maps for classifiers as described in Section \ref{related}, but none of them are suitable for detectors. We take SGLRP \cite{iwana2019explaining} as an example to introduce the relevance map for classifiers and then modify it for detectors because it excels in discriminating against irrelevant regions.

For a given deep classifier, the relevance map is defined as a normalized heat map with the same dimensions as input $x$, visualizing how $x$ contributes to a specific output node $t$ in pixel-wise manners. We denote this map as $h(x, t)$, which, in SGLRP, is obtained by back-propagating the ``relevance'' $R$ from the output layer (layer $L$) to the input layer (layer $1$) after the forward inference. Suppose layer $l$ has $N$ nodes (dimension of features) and layer $l+1$ has $M$ nodes, the relevance $R_n^{(l)}$ at node $n$ in layer $l$ is defined recursively by
\begin{equation*}
R_{n}^{(l)}=\sum_{m} \frac{a_{n}^{(l)} w_{n, m}^{+(l)}}{\sum_{n^{\prime}} a_{n^{\prime}}^{(l)} w_{n^{\prime}, m}^{+(l)}} R_{m}^{(l+1)},
\end{equation*}
for nodes with definite positive values (such as after ReLU), and
\begin{equation*}
R_{n}^{(l)}=\sum_{m} \frac{z_{n}^{(l)} w_{n, m}^{(l)}-b_{n}^{(l)} w_{n, m}^{+(l)}-h_{n}^{(l)} w_{n, m}^{-(l)}}{\sum_{n^{\prime}} z_{n^{\prime}}^{(l)} w_{n^{\prime}, m}^{(l)}-b_{n^{\prime}}^{(l)} w_{n^{\prime}, m}^{+(l)}-h_{n^{\prime}}^{(l)} w_{n^{\prime}, m}^{-(l)}} R_{m}^{(l+1)},
\end{equation*}
for nodes that may have negative values. In the formulas above, $a_n^{(l)}$ is the post-activation output of node $n$ in layer $l$ and $z_n^{(l)}$ is the pre-activation one. The range $[b_n^{(l)}, h_n^{(l)}]$ stands for the minimum and maximum of $z_n^{(l)}$, and $w_{n, m}^{+(l)}=\max \left(w_{n, m}^{(l)}, 0\right)$, $w_{n, m}^{-(l)}=\min \left(w_{n, m}^{(l)}, 0\right)$.

According to the relevance propagation rules above, the relevance map $h(x, t) = R^{(1)}$ is calculated by recursively back-propagating the relevance in the output layer $R^{(L)}$, of which the $n^\text{th}$ component $R_n^{(L)}$ is defined in SGLRP as
\begin{eqnarray}\label{sglrp}
R_n^{(L)}=\left\{\begin{array}{ll}
y_n\left(1-y_n\right) & n=t, \\
-y_n y_t & n \neq t,
\end{array}\right.
\end{eqnarray}
where $y_n$ is the predicted probability of class $n$, and $y_t$ is that for the single target class $t$. 

In detectors, however, we need the pixel-wise contributions from the input to $m$ bounding boxes. This multi-node relevance map could not be directly calculated by (\ref{sglrp}), so we naturally modify SGLRP as
\begin{eqnarray}\label{mn-sglrp}
R_{n}^{(L)}=\left\{\begin{array}{ll}
{y}_{n}\left(1-{y}_{n}\right) & n \in T, \\
- \frac{1}{m} {y}_{n} \sum_{i=1}^{m} {y}_{t_i}  & n \notin T,
\end{array}\right.
\end{eqnarray}
where ${y}_{t_i}$ is the predicted probabilities for one target output node $t_i$. $T$ is the set containing all target output nodes $\{t_1, t_2, ..., t_m\}$. With iNNvestigate Library \cite{alber2019innvestigate} to implement Multi-Node SGLRP and Deep Learning platforms supporting auto-gradient, the gradients from RAD loss $L_\text{RAD}(x) = h(x, T)$ to sample $x$ could be obtained according to the relevance propagation rules.

\begin{figure}[!htpb]
  \centering
  \includegraphics[width=\hsize]{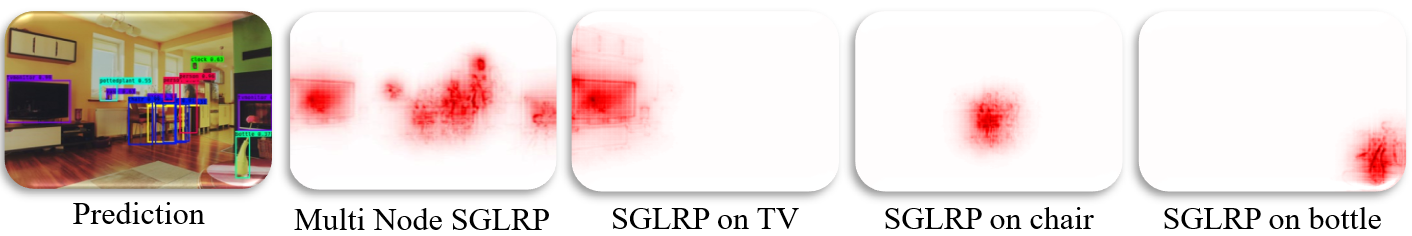}\\
  \caption{Difference between relevance maps from SGLRP and Multi-Node SGLRP. The relevance maps are for YOLOv3 \cite{redmon2018yolov3}.}
  \label{mnsglrp}
\end{figure}

We illustrate the difference between SGLRP and our Multi-Node SGLRP in Fig. \ref{mnsglrp}. SGLRP only displays the relevance map for one bounding box, e.g., ``TV'', ``chair'' and ``bottle''. Multi-Node SGLRP, in contrast, visualizes the overall relevance.

\subsection{Where to Attack?}\label{where}
Besides the calculation of relevance maps, it is also important to choose a proper node-set $T$ to attack. Specifically, we need to select certain bounding boxes and the corresponding output nodes for RAD.

Heuristically, the most ``obvious'' bounding boxes are desired to be eliminated, so we select the bounding boxes with the highest confidence, following \cite{xie2017adversarial}. Concretely, it is feasible to \emph{statically} choose several bounding boxes to attack in each iteration, or \emph{dynamically} attack all bounding boxes whose confidence exceeds a threshold. In our evaluation, the two strategies differ a little in performance and are not sensitive to hyper-parameter as demonstrated in \ref{strategy}. This shows that RAD does not require a sophisticated tuning of parameters, which is user-friendly. In our following experiments, we statically attack 20 nodes.

After selecting bounding boxes, we could attack their size, leading them to shrink; or their localization, leading them to shift; or their confidence, leading them to be misclassified. To adopt the best strategy, we conduct a toy experiment by attacking YOLOv3 \cite{redmon2018yolov3}, denoted as M2 (other models are specified in Table \ref{models}), following the settings later in Sec. \ref{expsec}. Given the results in Table \ref{strategytable}, the classification loss induces a better black-box transferability. This may be because detectors generally include a pre-trained classification as the feature extractor, and relevance maps are believed to be an indicator of successful attacks \cite{dong2019evading, zhang2019interpreting}. Note that our method is applicable to both one-stage detectors and two-stage ones because they both have classification outputs which we target on.

\begin{table}[!htpb]
  \caption{Detection mAP in RAD with different attacked nodes}
  \label{strategytable}
  \centering
  \begin{tabular}{r|ccccccccc}
\toprule
Strategy & M1 & M2 & M3 & M4 & M5 & M6 & M7 & M8 & M9\\ \hline
No Attack & 29.3  & 33.4  & 38.1  & 40.7  & 42.1  & 42.5  & 45.7  & 46.9 & 53.9\\
Size & 26.0  & 14.7  & 31.9  & 32.5  & 35.6  & 35.4  & 38.6  & 40.0 & 47.8 \\
Local. & 22.8  & 6.4  & 27.4  & 28.1  & 31.7  & 30.8  & 34.4  & 35.9 & 45.1 \\
Class. & \textbf{18.1}  & \textbf{1.2}  & \textbf{19.9}  & \textbf{20.5}  & \textbf{24.3}  & \textbf{22.6}  & \textbf{26.4}  & \textbf{28.2} & \textbf{39.9} \\
\bottomrule
  \end{tabular}
\end{table}

\subsection{How to Update?}\label{how}
With the relevance map $h(x, T)$ for certain attacked nodes $T$, we are able to attack, i.e., update the original sample to become adversarial by suppressing the relevance map according to the attack gradients $g(x)$ as
\begin{eqnarray}\label{grad}
g(x)=\frac{\partial L_\text{RAD}(x)}{\partial x}=\frac{\partial h(x, T)}{\partial x}.
\end{eqnarray}
Some update techniques are validated to be effective for enhancing the transferability in classification. For example, Scale-Invariant (SI) \cite{lin2019nesterov} proposes to average the attack gradients by scale copies of the samples as
\begin{eqnarray}\label{updatesi}
\begin{split}
g_\text{si}(x)= \frac{1}{k} \sum_{i=0}^{k} g(x/2^i).
\end{split}
\end{eqnarray}

Besides SI, Diverse Input (DI) \cite{xie2019improving}, Translation-Invariant (TI) \cite{dong2019evading} are also promising in classification. We are curious about whether they also work well in object detection. To explore this, we adopt these techniques in RAD as the setting suggested by their designers (see Sec. \ref{expsec} and \ref{detail}). From the results in Table \ref{technique}, we discover that SI is quite effective, further decreasing the mAP from the baseline significantly. Accordingly, RAD adopts (\ref{updatesi}) to update.

\begin{table}[!htpb]
  \caption{Detection mAP in RAD with different update techniques}
  \label{technique}
  \centering
  \begin{tabular}{r|ccccccccc}
\toprule
Technique & M1 & M2 & M3 & M4 & M5 & M6 & M7 & M8  & M9\\ \hline
None & 18.1  & 1.2  & 19.9  & 20.5  & 24.3  & 22.6  & 26.4  & 28.2 & 39.9 \\
DI & 18.1  & 1.0  & 19.9  & 20.5  & 23.9  & 22.4  & 26.3  & 27.9 & 39.6  \\
TI & 17.0  & 2.4  & 20.8  & 20.8  & 25.2  & 23.0  & 27.9  & 29.7 & 41.5 \\
SI & \textbf{14.6}  & \textbf{0.7}  & \textbf{16.3}  & \textbf{17.0}  & \textbf{20.4}  & \textbf{19.1}  & \textbf{22.3}  & \textbf{23.8} & \textbf{35.0} \\
\bottomrule
  \end{tabular}
\end{table}

With the calculated gradient, we update the sample like PGD attack \cite{madry2017towards} as
\begin{eqnarray}\label{update}
\begin{split}
x_{k+1} &= \text{clip}_\varepsilon\left(x_{k} - \alpha \frac{g_\text{si}(x_k)}{||g_\text{si}(x_k)||_1/N}\right),
\end{split}
\end{eqnarray}
where $\alpha$ stands for the step length. $x$ is $\ell_\infty$-norm bounded by $\varepsilon$ from the original sample in each iteration as in \cite{xie2019improving, lin2019nesterov, dong2019evading}. Gradient $g(x)$ is normalized by its average $\ell_1$-norm,i.e., $||g(x)||_1 /N$ to prevent numerical errors and control the degree of perturbations. $N$ is the dimension of the image, i.e., $N=height \times width \times channel$. Division by $N$ is necessary because $\ell_1$-norm sums all components of the tensor $x$, which is too large as a normalization factor. We do not adopt the mainstream sign method because it is not suitable to generate small perturbations as shown in other attacks in detectors \cite{xie2017adversarial}.

\section{Experiments}\label{expsec}
In this section, we evaluate the performance of RAD on various detectors, especially its transferability. The results are presented visually and numerically. In comprehensive evaluation, RAD achieves great transferability across models and even across tasks.

\subsection{Setup}
Our experiments are based on Keras \cite{chollet2015keras}, Tensorflow \cite{tensorflow2015-whitepaper} and PyTorch \cite{paszke2019pytorch} in 4 NVIDIA GeForce RTX 2080Ti GPUs. Library iNNvestigate \cite{alber2019innvestigate} is used to implement Multi-Node SGLRP.

We conduct experiments on MS COCO 2017 dataset \cite{lin2014microsoft}, which is a large-scale benchmark for object detection, instance segmentation, and image captioning. For a fair evaluation, we generate adversarial samples from all 5K samples in its validation set and test several black-box models on their mAP, a standard measure in many works \cite{he2017mask, chen2019hybrid}.

All attacks are conducted with the step length $\alpha=2$ for 10 iterations and the perturbation is $\ell_\infty$-bounded in $\varepsilon=16$ to guarantee the imperceptibility as in \cite{dong2019evading} if not particularly specified. To validate that the mAP drop comes from the attack instead of resizing or perturbation, we add large Gaussian noises ($\sigma=9$) to the resized images, and report it as ``Ablation''.

We choose 8 typical detectors ranging from the first end-to-end detector to state-of-the-art counterparts for attack and test. The variety of models guarantees the validity of results. We specify their information in Table \ref{models} and the corresponding pre-processing or details in \ref{detail}.

\begin{table}[!htpb]
\caption{Model backbone and mAPs}
\centering
\renewcommand\tabcolsep{4.5pt}
\label{models}
\begin{tabular}{ccccc}
  \toprule
ID & Model & Type & Backbone & mAP \\ \hline
M1 & SSD512 \cite{liu2016ssd} & one-stage & VGG16 & 29.3 \\
M2 & YOLOv3 \cite{redmon2018yolov3} & one-stage & Darknet & 33.4 \\
M3 & RetinaNet \cite{lin2017focal} & one-stage & ResNet-101 & 38.1  \\
M4 & Faster R-CNN \cite{ren2015faster} & two-stage & ResNeXt-101-64*4d & 40.7 \\
M5 & Mask R-CNN \cite{he2017mask} & two-stage &  ResNeXt-101-64*4d & 42.1 \\
M6 & Cascade RCNN \cite{cai2018cascade} & two-stage & ResNet-101 & 42.5 \\
M7 & Cascade Mask R-CNN \cite{cai2018cascade} & two-stage & ResNeXt-101-64*4d & 45.7  \\
M8 & Hyrbrid Task Cascade \cite{chen2019hybrid} & two-stage & ResNeXt-101-64*4d & 46.9  \\
M9 & EfficientDet \cite{tan2020efficientdet} & one-stage & EfficientNet + BiFPN & 53.9 \\
\bottomrule
\end{tabular}
\end{table}

\subsection{Visual Results of RAD}\label{expvisual}
\begin{figure}[!htpb]
  \centering
  \includegraphics[width=\hsize]{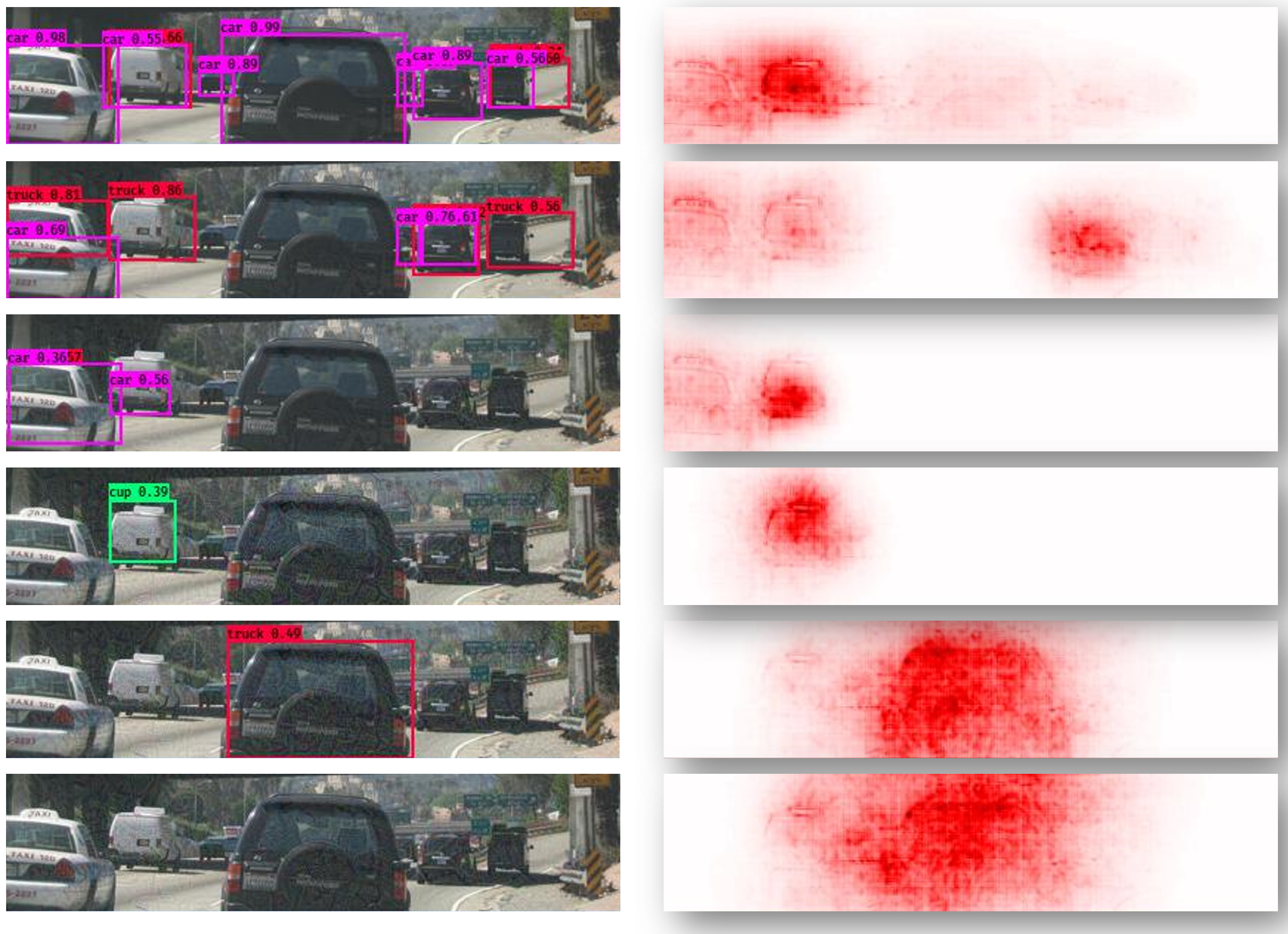}\\
  \caption{The model predictions and the relevance map in different attack iterations during RAD on YOLOv3. We could see from top to bottom how RAD gradually influences the prediction of YOLOv3 (left column) by distorting its relevance map (right column).}
  \label{convergence}
\end{figure}

\begin{figure}[!htpb]
  \centering
  \includegraphics[width=\hsize]{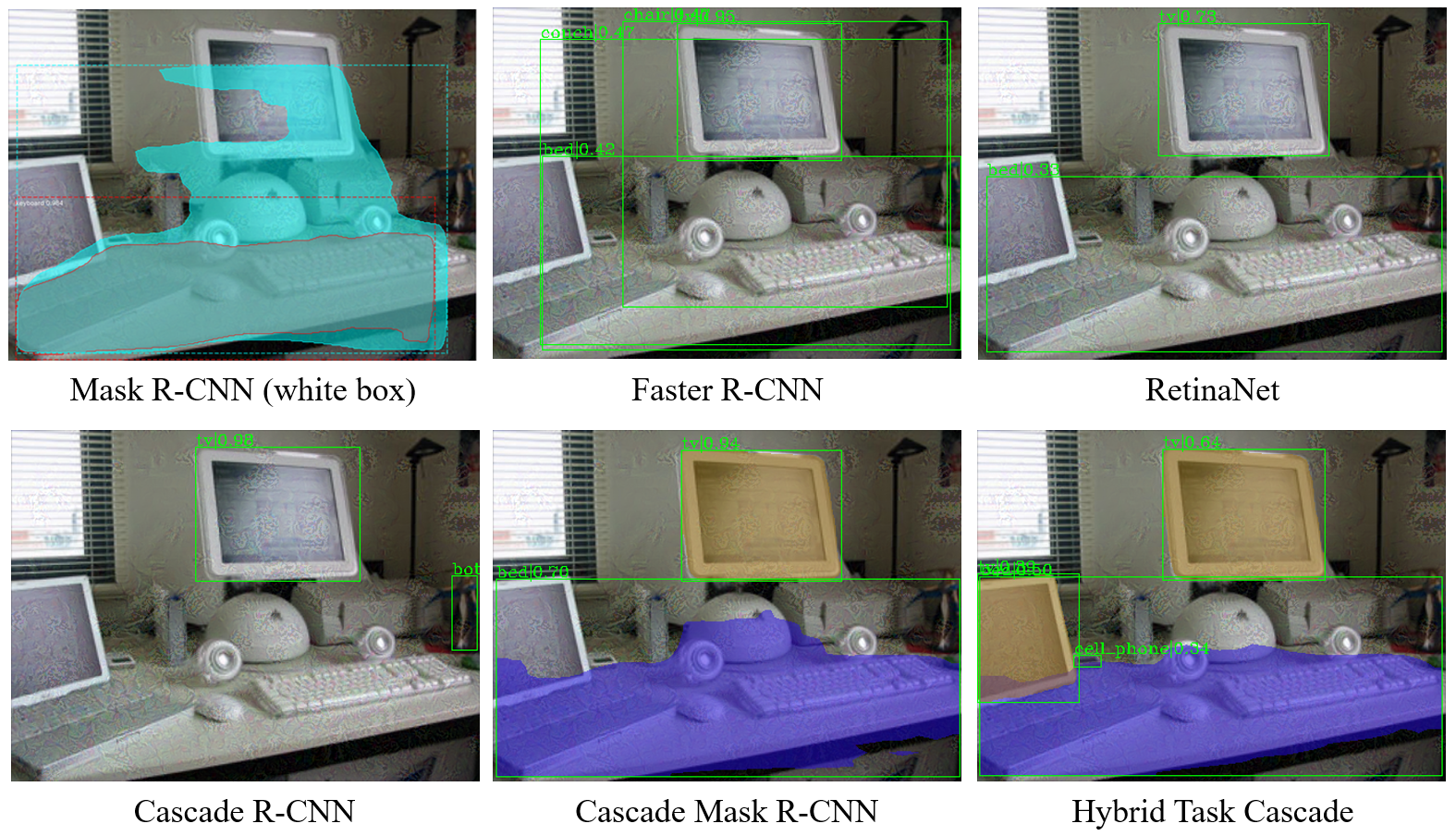}\\
  \caption{RAD has a great transferability. The same adversarial sample generated by attacking Mask R-CNN fools all 5 black-box detectors.}
  \label{black}
\end{figure}

We first intuitively illustrate the attack process in Fig. \ref{convergence} and the attack transferability in Fig. \ref{black}.

By RAD, the relevance map is attacked to be meaningless and loses its focus. In Fig. \ref{convergence}, the initial prediction is correct and the relevance map is clear. RAD constantly misleads the relevance map to be unstructured without the outline of objects. Finally, all bounding boxes vanish.

In Fig. \ref{black}, we visualize several predictions on the same adversarial sample by black-box models. The objects in the image, e.g., the laptop and keyboard, are quite large and obvious to detect. However, with a small perturbation from RAD, 5 black-box models all fail to detect the laptop, keyboard, and mouse. Surprisingly, 4 of them even detect a non-existent ``bed'', which is neither relevant nor similar in the image.

\subsection{RAD's Transferability in Object Detection}\label{transcompare}
To evaluate the in-domain transferability of detection attacks and cross-domain transferability of classification attacks, we test the detection mAP of 8 models in COCO adversarial samples generated in the setting stated before.

For detection attacks, adversarial samples are crafted by attacking surrogate model M2 (YOLOv3 \cite{redmon2018yolov3}). For classification attacks, we use the model output on the clean sample as the label. By several state-of-the-art attacks on surrogate classifiers (InceptionV3 \cite{szegedy2016rethinking} here as in \cite{xie2019improving, dong2019evading}), the adversarial samples are generated and tested the mAP as the transferability towards detectors. Details of implementation are described in \ref{detail}.

We present the results in Table \ref{trans}. Among the classification attacks and detection ones, cross-domain attack \cite{naseer2019cross} is effective, but RAD is more aggressive. RAD enjoys a state-of-the-art transferability towards most black-box models, outperforming other methods for above 20\%. The detection mAPs are more than halved, making state-of-the-art detectors similar to the early elementary counterpart. Also, the adversarial samples from the one-stage detector (M1) could be transferred to two-stage ones (M4 - M8, seeing Table \ref{models}) as the case in attacking classifiers \cite{dong2018boosting, xie2019improving, lin2019nesterov}.

\begin{table}[!htpb]
  \caption{Detection mAP in different attacks}
  \label{trans}
  \centering
  \renewcommand\tabcolsep{4pt}
  \begin{tabular}{r|ccccccccc}
\toprule
 Method & M1 & M2 & M3 & M4 & M5 & M6 & M7 & M8 & M9 \\ \hline
No Attack & 29.3  & 33.4  & 38.1  & 40.7  & 42.1  & 42.5  & 45.7  & 46.9 & 53.9 \\
Ablation & 24.9  & 31.4  & 31.2  & 31.6  & 35.0  & 34.3  & 37.5  & 38.8 & 48.6 \\ \midrule

PGD \cite{madry2017towards} & 26.4  & 30.4  & 34.4  & 35.4  & 38.4  & 38.3  & 41.7  & 43.1 & 51.1 \\
SI-PGD \cite{lin2019nesterov} & 27.5 & 31.6  & 36.1  & 37.1  & 40.0  & 40.1  & 43.5  & 44.8 & 52.4 \\
MI-DI-PGD \cite{dong2018boosting, xie2019improving} & 22.9  & 26.2  & 29.3  & 30.0 & 33.2  & 32.1  & 36.0  & 37.5 & 48.0 \\
MI-TI-PGD \cite{dong2018boosting, dong2019evading} & 20.1  & 23.7  & 24.9  & 25.4  & 30.1  & 27.4  & 32.8  & 34.5 & 47.1 \\
CD-painting \cite{naseer2019cross} & 16.4  & 20.8  & 21.3  & 22.8  & 26.6  & 24.5  & 28.9  & 29.5 & 42.3 \\
CD-comics \cite{naseer2019cross} & 16.6  & 21.6  & 21.7  & 22.7  & 26.8  & 24.3  & 29.1  & 42.3 & 43.7 \\ \midrule

Dfool \cite{lu2017adversarial} & 23.3  & 2.5  & 29.2  & 29.8  & 33.3  & 32.9  & 36.5  & 38.0 & 47.5 \\
Loc \cite{zhang2019towards} & 21.9  & \textbf{0.2}  & 25.8  & 26.6  & 29.8  & 29.4  & 33.2  & 33.2 & 45.2 \\
DAG \cite{xie2017adversarial}& 20.8  & 0.6  & 22.8  & 23.4  & 26.8  & 25.6  & 28.9  & 31.0 & 40.6 \\ 
RAD (ours)& \textbf{14.6}  & 0.7  & \textbf{16.3}  & \textbf{17.0}  & \textbf{20.4}  & \textbf{19.1}  & \textbf{22.3}  & \textbf{23.8} & \textbf{35.0} \\
\bottomrule
  \end{tabular}
\end{table}

\begin{figure}[!htpb]
  \centering
  \includegraphics[width=\hsize]{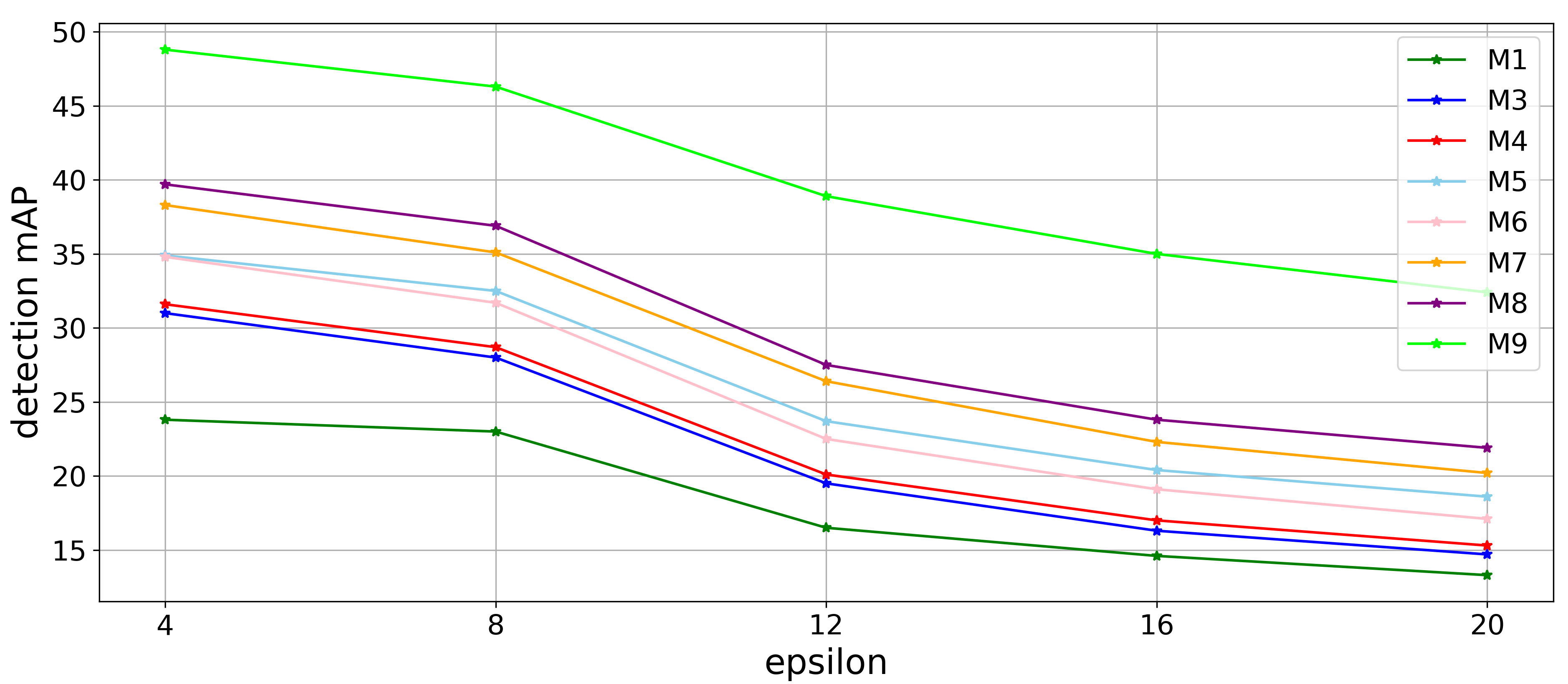}\\
  \caption{The influence of $\varepsilon$ on detection mAP in RAD}
  \label{eps}
\end{figure}

\subsection{Comprehensive Evaluations of RAD's Transferability}\label{investigate}
Although mAP is a general metric to test detectors under a fixed bound of perturbations, it is also interesting to investigate RAD's transferability under different bounds and sub-tasks, i.e., classification accuracy, the shift of bounding boxes, and their invisibility to detectors.

Here we first vary the $\ell_\infty$ bound of RAD, and report the mAP in Fig. \ref{eps}. With the $\ell_\infty$ bound increases, the resulting mAP greatly decreases for all black-box models especially for $\varepsilon$ from 8 to 12.

We further study RAD's transferability in different metrics, which is necessary because mAP measures the overall performance of detectors, and thus could not decouple the RAD's aggression on, e.g., shifting and hiding bounding boxes. For subsequent evaluations on sub-tasks, we define the metrics below.
\begin{itemize}
  \item \textbf{Accuracy} for \textbf{classification} of bounding boxes: Without considering locations, we use the predicted box classes to calculate the classification accuracy. For each image, $a$ predicted bounding boxes with the highest confidence (after non-maximum suppression) are considered, where $a$ is the number of ground truth bounding boxes for rationality. For example, if the detector predicts 3 cats, 2 dogs, and 1 car in an image, which has 2 cats, 3 dogs, and 1 person, we count $2+2=4$ hits and 2 misses.
  \item \textbf{Intersection over Union (IoU)} for \textbf{shifting} of bounding boxes: Focused on locations, we average the IoU, the measure of location correctness, on all predictions to quantify the shifting. Also, we select $a$ predicted bounding boxes with the highest confidence, and compute the IoU for each box with the nearest ground truth in the same class (IoU=0 for wrong prediction).
  \item \textbf{Mean Average Recall (mAR)} for \textbf{hiding} of bounding boxes: To see how attacks hide the object, the common recall value is appropriate, and the mAR is calculated from recall in the same way that mAP is from precision, which has been adopted in detection libraries.
\end{itemize}
To the best of our knowledge, existing works mostly focus on mAP \cite{li2018robust, li2018exploring, liu2018dpatch}, and we here suggest a more comprehensive way to evaluate attacks on detectors. The results are reported in Table \ref{bboxrc}, from which one could observe that RAD also outperforms its counterparts in different sub-tasks to a large extent, i.e., up to 10\%. The predicted bounding boxes in attacking white-box M2 are too few to evaluate fairly, so we do not show their results.

\begin{table}[!htpb]
  \caption{Accuracy (\%), IoU (\%), mAR (\%) in different attacks}
  \label{bboxrc}
  \centering
  \renewcommand\tabcolsep{5pt}
  \begin{tabular}{c|r|cccccccc}
\toprule
Metric & Method & M1 & M3 & M4 & M5 & M6 & M7 & M8 & M9 \\ \hline
\multirow{4}{*}{Acc}
& Dfool \cite{lu2017adversarial} & 68.9   & 71.7  & 72.6  & 74.8  & 73.5  & 75.9  & 76.6 & 82.9 \\
& Loc \cite{zhang2019towards} & 67.8   & 69.8  & 70.7  & 73.0  & 71.5  & 74.0  & 74.8 & 81.6 \\
& DAG \cite{xie2017adversarial}& 64.4   & 62.4  &63.3  & 67.2  & 65.0  & 68.5  & 69.3 & 77.1 \\
& RAD (ours) & \textbf{58.6}  & \textbf{56.9}  & \textbf{58.0}  &\textbf{62.2}  & \textbf{60.3}  & \textbf{63.5}  & \textbf{64.2} & \textbf{72.9} \\ \midrule 
\multirow{4}{*}{IoU}
& Dfool \cite{lu2017adversarial} & 43.9   & 47.9  & 48.5  & 51.3  & 50.1  & 52.6  & 53.1 & 61.5 \\
& Loc \cite{zhang2019towards} & 42.2   & 44.6  & 45.1  & 47.9  & 46.7  & 49.3  & 49.8 & 59.6 \\
& DAG \cite{xie2017adversarial}& 40.9   & 41.1  & 41.7  & 45.3  & 43.7  & 46.8  & 46.9 & 56.3 \\
& RAD (ours) & \textbf{34.4}  & \textbf{34.5}  & \textbf{35.6}  &\textbf{39.5}  & \textbf{37.9}  & \textbf{41.1}  & \textbf{40.9} & \textbf{51.7} \\ \midrule 
\multirow{4}{*}{mAR}
& Dfool \cite{lu2017adversarial} & 36.3   & 45.9  & 43.9  & 47.1  & 45.9  & 48.8  & 54.8 & 60.7 \\
& Loc \cite{zhang2019towards} & 34.7   & 42.3  & 40.5  & 43.7  & 42.5  & 45.4  & 51.1 & 58.3 \\
& DAG \cite{xie2017adversarial}& 32.9   & 38.5  & 35.3  & 39.2  & 36.7  & 39.4  & 47.4 & 52.8 \\
& RAD (ours) & \textbf{27.4}  & \textbf{32.2}  & \textbf{29.0}  &\textbf{32.7}  & \textbf{30.1}  & \textbf{32.9}  & \textbf{41.6} & \textbf{47.9} \\ 
\bottomrule
  \end{tabular}
\end{table}

\subsection{RAD's Transferability to Instance Segmentation}
Detection and segmentation are similar in some aspects, so they could be implemented in one network \cite{he2017mask, cai2018cascade, chen2019hybrid}. Also, adversarial samples for object detection tend to transfer to instance segmentation \cite{xie2017adversarial}. Accordingly, we evaluate this cross-task transferability by RAD on surrogate detectors YOLOv3 (\cite{redmon2018yolov3}, M2), RetinaNet (\cite{lin2017focal}, M3) and Mask R-CNN (\cite{he2017mask}, M5). From the results in Table \ref{segm}, we find that RAD also greatly hurts the performance of instance segmentation, leading to a drop in mAP of over 70\%. This inspires the segmentation attackers to indirectly attack detectors.
\begin{table}[!htpb]
  \caption{Segmentation mAP of RAD}
  \label{segm}
  \centering
  \begin{tabular}{r|ccc|ccc|ccc}
\toprule
 & \multicolumn{3}{c}{mAP} & \multicolumn{3}{c}{mAP50} & \multicolumn{3}{c}{mAP75} \\
Surrogate & M5 & M7 & M8 & M5 & M7 & M8 & M5 & M7 & M8 \\ \hline
None & 38.0  & 39.4  & 40.8  & 60.6  & 61.3  & 63.3  & 40.9  & 42.9  & 44.1  \\
Ablation & 31.0  & 31.9  & 33.5  & 51.2  & 51.0  & 53.7  & 32.4  & 34.3  & 35.4  \\ \midrule
M2 & 17.9  & 18.6  & 20.3  & 31.6  & 31.7  & 34.5  & 18.0  & 18.9  & 20.7  \\
M3 & 11.6  & 11.9  & 12.9  & 19.2  & 19.1  & 20.7  & 12.1  & 12.6  & 13.7  \\
M5 & 1.2  & 11.1  & 11.8  & 2.4  & 17.9  & 18.9  & 1.0  & 11.9  & 12.6  \\
\bottomrule
  \end{tabular}
\end{table}

\section{Adversarial Objects in Context}
Given the great transferability of RAD, we create Adversarial Objects in COntext (AOCO), the first adversarial dataset for object detection and instance segmentation. AOCO dataset serves as a potential benchmark to evaluate the robustness of detectors, which is beneficial to network designers. It will also be useful for adversarial training, as the most effective practice to improve the robustness of DNNs. Notice that there is no other adversarial dataset for detection and segmentation at all. This is not because the dataset is useless, but due to the low transferability of attack methods such that the examples are detector-dependent. Now we have achieved high transferability and can then make such an adversarial dataset publicly available.

AOCO is generated from the full COCO 2017 validation set \cite{lin2014microsoft} with 5k samples. It contains 5K adversarial samples for evaluating object detection (AOCO detection) and 5K for instance segmentation (AOCO segmentation). All 10K samples in AOCO are crafted by RAD. The surrogate model we attack is YOLOv3 for AOCO detection and Mask R-CNN for AOCO segmentation given the results in Table \ref{trans} and Table \ref{segm}.

\begin{figure}[!htpb]
  \centering
  \includegraphics[width=\hsize]{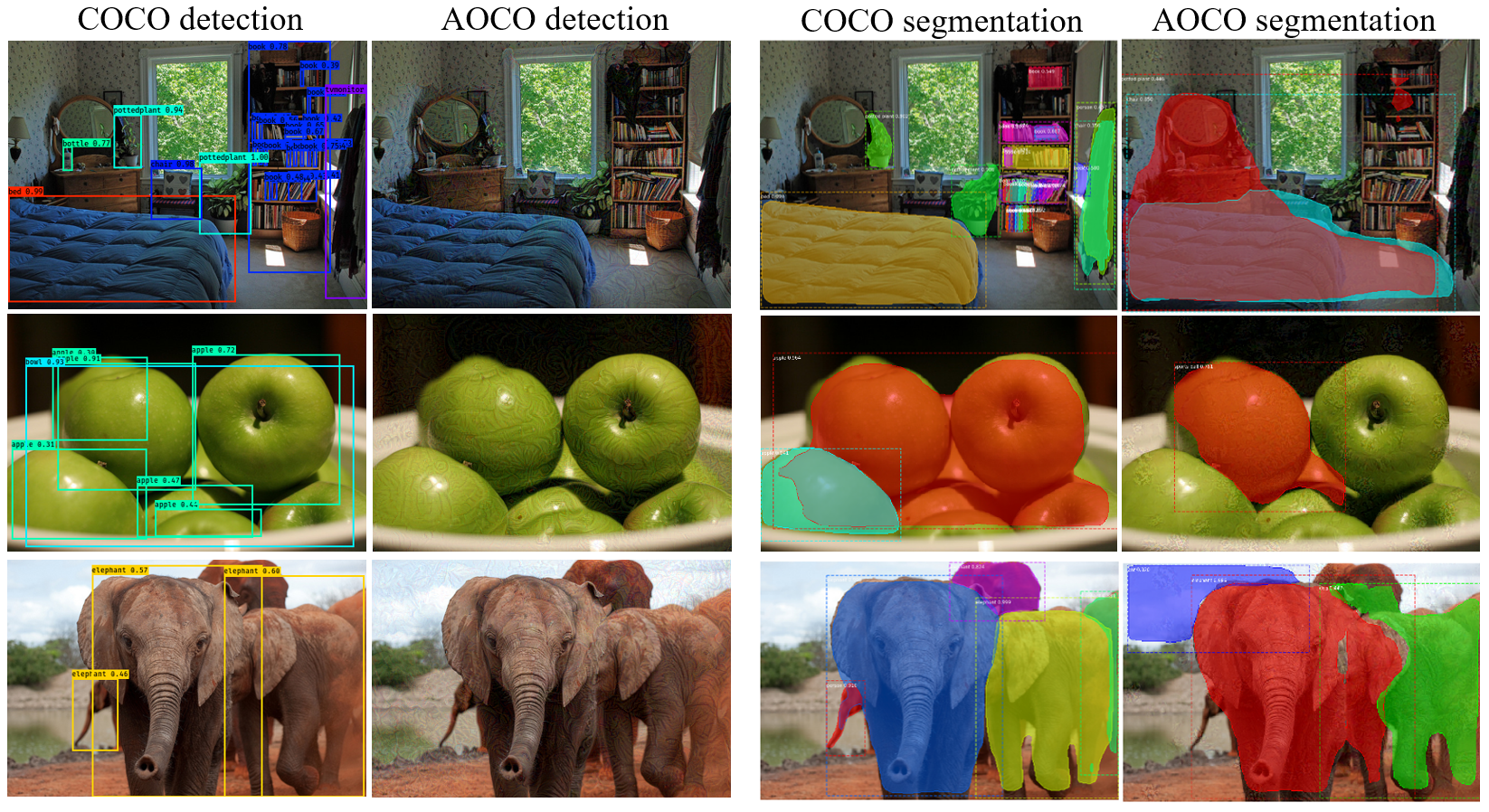}\\
  \caption{Detection and segmentation results in COCO and AOCO by YOLOv3 and Mask R-CNN. For COCO, both networks predict correctly. For AOCO segmentation results, the top image contains two big masks for ``chair'' and ``potted plan''; the second image contains one false mask for ``sports ball''; the bottom image contains ``dog'' in green, ``car'' in purple and ``elephant'' in red.}
  \label{aocoappendix}
\end{figure}

We measure the perturbation $\Delta x$ in AOCO by Root Mean Squared Error (RMSE) as in \cite{xie2017adversarial}. It is calculated as $\sqrt{\sum_i (\Delta x_i)^{2} / N}$ in a pixel-wise way, and $N$ is the size of the image. Performance of AOCO is reported in Table \ref{aoco}. The RMSE in AOCO is 6.469 for detection and 6.606 for segmentation, which is lower than that in \cite{wu2019skip}, and the perturbations are quite imperceptible. We show in Fig. \ref{aocoappendix} the AOCO samples. More details are presented in \ref{moreaoco}.
\begin{table}[!htpb]
  \caption{Detection mAP and segmentation mAP on COCO and AOCO}
  \label{aoco}
  \centering
  \begin{tabular}{l|ccccccccc}
\toprule
 &  M1 & M2 & M3 & M4 & M5 & M6 & M7 & M8 & M9 \\ \hline
COCO det. & 29.3  & 33.4  & 38.1  & 40.7  & 42.1 & 42.5  & 45.7  & 46.9 & 53.9 \\
AOCO det. & 14.6 & 0.7 & 16.3 & 17 & 20.4 & 19.1 & 22.3 & 23.8 & 35.0\\ \midrule
COCO seg. & $\backslash$ & $\backslash$ & $\backslash$ & $\backslash$ & 38.0 & $\backslash$ & 39.4 & 40.8 & $\backslash$\\
AOCO seg. & $\backslash$ & $\backslash$ & $\backslash$ & $\backslash$ & 1.2 & $\backslash$ & 11.1 & 11.8 & $\backslash$\\
\bottomrule
  \end{tabular}
\end{table}

\section{Conclusion and Future Work}
To pursue a high transferability, this paper proposes Relevance Attack on Detectors (RAD), which works by suppressing the multi-node relevance, a common property across detectors calculated by our Multi-Node SGLRP. We also thoroughly discuss where to attack and how to update in attacking relevance maps. RAD achieves a state-of-the-art transferability towards 8 diverse black-box models, exceeding existing results by above 20\%, and also significantly hurts the instance segmentation. Given the great transferability of RAD, we generate the first adversarial dataset for object detection and instance segmentation, i.e., Adversarial Objects in COntext (AOCO), which helps to quickly evaluate and improve the robustness of detectors.

In the future, it is promising to attack other common properties for good transferability. Besides, RAD could be modified for patch-based attacks by releasing the bound of perturbations and focusing them only on the regions with the largest perturbations.

\section*{Acknowledgments}
The authors are grateful to the anonymous reviewers for their insightful comments. This work was partially supported by National Key Research Development Project (No. 2018AAA0100702), National Natural Science Foundation of China (No. 61977046), 1000-Talent Plan (Young Program), and Shanghai Municipal Science and Technology Major Project (2021SHZDZX0102).

\newpage
\bibliography{Reference}

\newpage
\appendix

\section{Influence of Hyper-Parameters in Node Selection}\label{strategy}
Performance of RAD is not sensitive to hyper-parameter, no matter the strategy to select bounding boxes is dynamic or static as Table \ref{hyper}. Attackers are not bothered to tune them carefully. The parameter for dynamic strategy refers to the pre-softmax confidence threshold to select a bounding box. The parameter for static strategy refers to the fixed number of selected bounding boxes in each iteration.
\begin{table}[!htpb]
  \caption{Detection mAP in different hyper-parameters in RAD}
  \label{hyper}
  \centering
  \begin{tabular}{c|c|ccccccccc}
\toprule
Str. & Para. & M1 & M2 & M3 & M4 & M5 & M6 & M7 & M8 & M9\\ \hline
\multirow{3}{*}{Dyn.} & -1 & 18.3  & 1.2  & 20.0  & 20.6  & 24.3  & 22.8  & 26.7  & 28.3 & 40.0\\
 & -2 & 18.2  & 1.2  & 20.1  & 20.7  & 24.2  & 22.8  & 26.6  & 28.0 & 39.8\\
& -3 & 18.4  & 1.3  & 20.3  & 20.8  & 24.3  & 22.9  & 26.7  & 28.5 & 40.2\\ \midrule

\multirow{3}{*}{Sta.} & 10 & 18.2  & 1.1  & 19.9  & 20.5  & 24.2  & 22.8  & 26.2  & 28.0 & 39.8\\
 & 20 & 18.1 & 1.2  & 19.9  & 20.5  & 24.3  & 22.6  & 26.4  & 28.2 & 39.9 \\
& 30 & 18.2  & 1.3  & 20.1  & 20.7  & 24.3  & 22.8  & 26.0  & 27.9 & 40.1\\
\bottomrule
  \end{tabular}
\end{table}

\section{RAD on More Surrogates}\label{surrogate}
The results on attacking more surrogates by RAD are reported in Table \ref{mapsurrogate}.
\begin{table}[!htpb]
  \caption{Detection mAP of RAD on different surrogates}
  \label{mapsurrogate}
  \centering
  \renewcommand\tabcolsep{5pt}
  \begin{tabular}{r|ccccccccc}
\toprule
Surrogate & M1 & M2 & M3 & M4 & M5 & M6 & M7 & M8 & M9 \\ \hline
M2 (YOLOv3) & 14.6  & 0.7  & 16.3  & 17.0  & 20.4  & 19.1  & 22.3  & 23.8 & 35.0 \\
M3 (RetinaNet) & 20.7  & 6.1  & 2.3  & 25.7 & 29.3  & 28.2  & 31.7  & 33.8  & 44.1 \\
M5 (Mask R-CNN) & 20.4 & 24.2 & 25.7 & 26.5 & 1.1 & 28.9  & 33.1  & 34.9 & 45.4 \\
\bottomrule
  \end{tabular}
\end{table}

\section{Implementation Details}\label{detail}
\subsection{Pre-processing}
To pre-process, we resize the image with its long side as 416 for YOLOv3 or RetinaNet and 448 for Mask R-CNN, and then zero-pad it to a square. The resolution is kept relatively the same for a fair evaluation. Images are normalized to [0,1] in YOLOv3 or subtracted by the mean of COCO training set in RetinaNet and Mask R-CNN. Accordingly, samples in AOCO detection have the long side 416, and that for AOCO segmentation is 448.

\subsection{Transfer-Enhancing Update Techniques}
DI \cite{xie2019improving} transforms the image for 4 times with probability $p$ ($p=1$ for better transferability as suggested) and averaging the gradients. The transformation is to resize the image to $0.9\times$ its size and randomly padding the outer areas with white pixels. SI \cite{lin2019nesterov} divides the sample numerically by the power 2 for 4 times and averages the 4 obtained gradients. TI \cite{dong2019evading} translates the image to calculate the augmented gradients. To implement it efficiently, it adopts a kernel to simulate the averaging of gradients. We choose the kernel size 15 as suggested. MI \cite{dong2018boosting} uses momentum optimization (parameter $\mu=1$ as suggested) for a better transferability and a faster attack. Cross-domain attack \cite{naseer2019cross} uses extra datasets (paintings, denoted as CD-paintings, and comics, denoted as CD-comics) to train a perturbation generator with the relative loss. The adopted surrogate model is also InceptionV3 for consistency. Perturbations are resized to fit the sample size.

\subsection{Detection Attacks}
For DAG \cite{xie2017adversarial}, we follow the setting of generating dense proposals. The classification probabilities of 3000 bounding boxes with the highest confidence are attacked. But we alter its optimization to (\ref{update}) because its original update produces quite a small perturbation, leading to a poor transferability, which is unfair for comparison. Dfool \cite{lu2017adversarial} suppresses the classification confidence for the original bounding boxes, which is the same in our experiment. Localization loss is shown to be useful in \cite{zhang2019towards}, and here we suppress the width and height of the original bounding boxes.

\section{More about AOCO}\label{moreaoco}
We report the mAP50 and mAP75 of AOCO in Table \ref{aoco50} and Table \ref{aoco75}.
\begin{table}[!htpb]
  \caption{Detection mAP50 and segmentation mAP50 on COCO and AOCO}
  \label{aoco50}
  \centering
  \begin{tabular}{l|ccccccccc}
\toprule
 & M1 & M2 & M3 & M4 & M5 & M6 & M7 & M8 & M9\\ \hline
COCO det. & 49.2  & 56.4  & 58.1  & 62.0  & 63.8  & 60.7  & 64.1  & 66.0 & 74.3 \\
AOCO det. & 26.7  & 1.6  & 27.6  & 29.1  & 34.5  & 29.9  & 34.3  & 37.4 & 51.7 \\ \midrule
COCO seg. & $\backslash$ & $\backslash$ & $\backslash$ & $\backslash$ & 60.6  & $\backslash$ & 61.3  & 63.3 & $\backslash$ \\
AOCO seg. & $\backslash$ & $\backslash$ & $\backslash$ & $\backslash$ & 2.4  & $\backslash$ & 17.9  & 18.9 & $\backslash$ \\
\bottomrule
  \end{tabular}
\end{table}

\begin{table}[!htpb]
  \caption{Detection mAP75 and segmentation mAP75 on COCO and AOCO}
  \label{aoco75}
  \centering
  \begin{tabular}{l|ccccccccc}
\toprule
 & M1 & M2 & M3 & M4 & M5 & M6 & M7 & M8 & M9 \\ \hline
COCO det. & 30.8  & 35.8  & 40.6  & 44.6  & 46.3  & 46.3  & 50.0  & 51.2 & 59.9 \\
AOCO det. & 14.2  & 0.6  & 16.5  & 17.1  & 20.8  & 19.9  & 23.3  & 24.6 & 37.4 \\ \midrule
COCO seg. & $\backslash$ & $\backslash$ & $\backslash$ & $\backslash$ & 40.9  & $\backslash$ & 42.9  & 44.1 & $\backslash$ \\
AOCO seg. & $\backslash$ & $\backslash$ & $\backslash$ & $\backslash$ & 1.0  & $\backslash$ & 11.9  & 12.6 & $\backslash$ \\
\bottomrule
  \end{tabular}
\end{table}

\section{More Visualizations of the Relevance Maps}\label{visualrlv}
\begin{figure}[!htpb]
  \centering
  \includegraphics[width=\hsize]{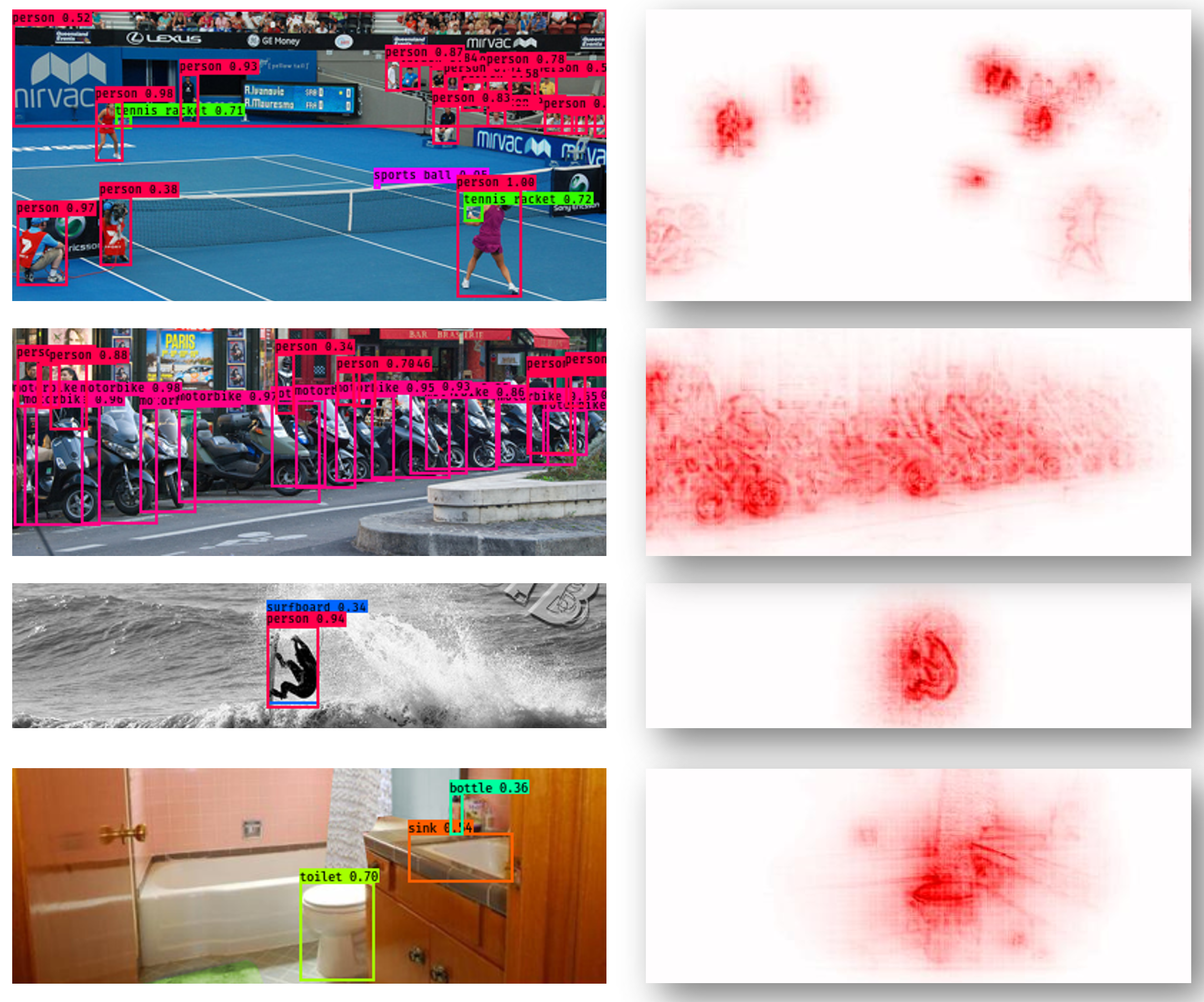}\\
  \caption{The detection results (left) and the corresponding relevance maps (right) by Multi-Node SGLRP on YOLOv3.}
  \label{morevis1}
\end{figure}

\begin{figure}[!htpb]
  \centering
  \includegraphics[width=\hsize]{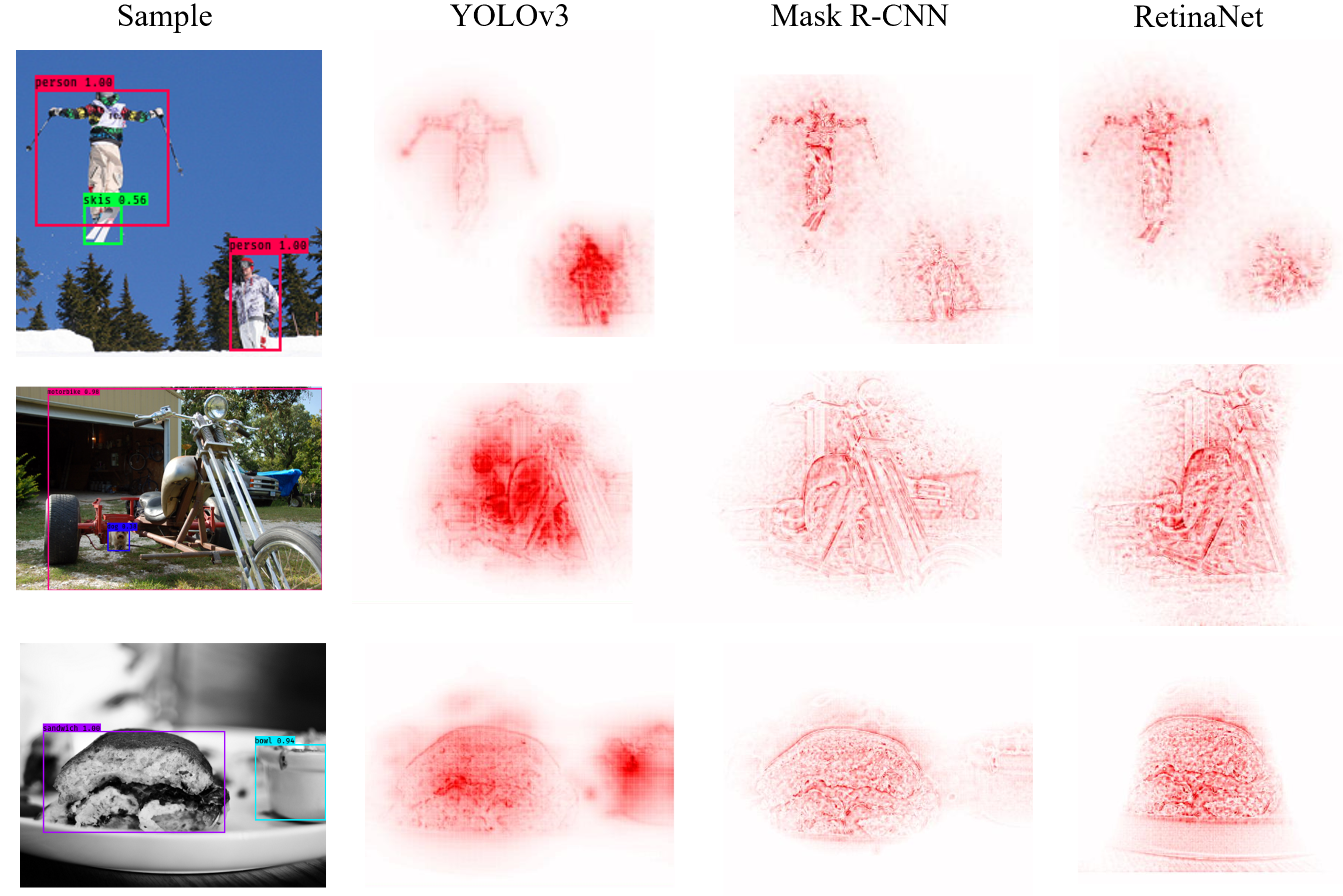}\\
  \caption{The detection results (leftmost) and the corresponding relevance maps by Multi-Node SGLRP on different detectors.}
  \label{morevis2}
\end{figure}

We present more visualizations to illustrate that the relevance map highlights the salient part for detection as in Fig. \ref{morevis1}, and it shares similarities across different models as in Fig. \ref{morevis2}.

\end{document}